%% file: main.tex
\documentclass[]{style/ceurart}
\usepackage{listings}
\begin{document}

\copyrightyear{2025}
\copyrightclause{Copyright for this paper by its authors.
  Use permitted under Creative Commons License Attribution 4.0
  International (CC BY 4.0).}

\conference{CLEF 2025 Working Notes, 9 -- 12 September 2025, Madrid, Spain}

\title{Hallucination Detection and Mitigation in Scientific Text Simplification using Ensemble Approaches: DS@GT at CLEF 2025 SimpleText}

\author[1]{Krishna Chaitanya Marturi}[
    email=kmarturi3@gatech.edu,
]
\cormark[1]
\author[2]{Heba H. Elwazzan}[
    email=helwazzan3@gatech.edu
]

\address[1]{Georgia Institute of Technology, North Ave NW, Atlanta, GA 30332}

\cortext[1]{Corresponding author.}

\begin{abstract}
    In this paper, we describe our methodology for the CLEF 2025 SimpleText Task 2, which focuses on detecting and evaluating creative generation and information distortion in scientific text simplification. Our solution integrates multiple strategies: we construct an ensemble framework that leverages BERT-based classifier, semantic similarity measure, natural language inference model, and large language model (LLM) reasoning. These diverse signals are combined using meta-classifiers to enhance the robustness of spurious and distortion detection. Additionally, for grounded generation, we employ an LLM-based post-editing system that revises simplifications based on the original input texts.

\end{abstract}


\maketitle

\input{sections/00_main}

\bibliography{main}

\appendix

\section{Prompt Templates}

\subsection{LLM as Judge Prompt}
\label{app:prompt-LLM-as-Judge}

\begin{lstlisting}
You are an expert annotator tasked with evaluating whether an input text is spurious when compared to a source document.

An input text is **spurious** if:
- It fabricates information not grounded in any source.
- It misrepresents or contradicts the source documents.
- It is too general, trivial, or irrelevant in the context of the documents, even if technically true.

Please review the following examples to guide your evaluation:

Example 1:
SOURCE DOCUMENT:
Online social media provide users with opportunities to engage with diverse opinions and can spread misinformation.

INPUT TEXT:
Social media always spreads misinformation.

RESPONSE:
{
  "spuriousness": 0.8,
  "over_generalization": 0.9,
  "contradiction": 0.7,
  "vagueness": 0.4
}

Example 2:
SOURCE DOCUMENT:
We propose a new welfare criterion that allows us to rank alternative financial market structures in the presence of belief heterogeneity.

INPUT TEXT:
We propose a new economic theory to manage inflation.

RESPONSE:
{
  "spuriousness": 1.0,
  "over_generalization": 0.8,
  "contradiction": 0.6,
  "vagueness": 0.5
}

Example 3:
SOURCE DOCUMENT:
We analyze economies with complete and incomplete financial markets and restricted trading possibilities like borrowing limits.

INPUT TEXT:
We analyze economies with complete and incomplete financial markets.

RESPONSE:
{
  "spuriousness": 0.1,
  "over_generalization": 0.4,
  "contradiction": 0.0,
  "vagueness": 0.2
}

Now evaluate the next pair:

SOURCE DOCUMENT:
{source}

INPUT TEXT:
{input_text}

Please answer with a score between 0 and 1 for each of the following:

1. Spuriousness (fabricated, irrelevant, or ungrounded): 
2. Over-generalization (too broad or omits key details): 
3. Contradiction (misrepresents or opposes the source): 
4. Vagueness (too imprecise, lacks specificity):

Do not include any text or commentary outside of the JSON response format below.
Respond in this JSON format:
{
  "spuriousness": float,
  "over_generalization": float,
  "contradiction": float,
  "vagueness": float
}
\end{lstlisting}

\subsection{LLM-based Classification Prompt}
\label{app:prompt-LLM Classifier}
\begin{lstlisting}
You are an expert linguist specializing in text simplification. Given a source sentence and a simplified version, your task is to score between 0 and 1 for any of the information distortion errors in the simplified sentence.

The possible error categories are:
- 'No error': 1 if the simplified sentence is accurate, grammatical, and faithful to the source.
- 'A1. Random generation': 1 if the simplified sentence contains unrelated content.
- 'A2. Syntax error': 1 if the simplified sentence is syntactically broken or malformed.
- 'A3. Contradiction': 1 if the simplified sentence contradicts the source.
- 'A4. Simple punctuation / grammar errors': 1 if there are small grammar or punctuation issues.
- 'A5. Redundancy': 1 if the sentence repeats information unnecessarily.
- 'B1. Format misalignement': 1 if the formatting deviates in a way that distorts meaning.
- 'B2. Prompt misalignement': 1 if the simplification ignores the intent or task of simplification.
- 'C1. Factuality hallucination': 1 if the simplified sentence adds false information.
- 'C2. Faithfulness hallucination': 1 if information is not supported by the source.
- 'C3. Topic shift': 1 if the sentence shifts to a different topic.
- 'D1.1. Overgeneralization': 1 if it makes the content too broad and less precise.
- 'D1.2 Overspecification of Concepts': 1 if it introduces more specificity than appropriate.
- 'D2.1. Loss of Informative Content': 1 if key details from the source are missing.
- 'D2.2. Out-of-Scope Generation': 1 if the simplified sentence includes irrelevant elaborations.

Please review the following example to guide your evaluation:

Example 1:
SOURCE SENTENCE: We conducted the experiments with a data set of 94 chest CTs (laboratory confirmed 39 viral bronchiolitis caused by human parainfluenza (HPIV), 34 nontuberculous mycobacterial (NTM), and 21 normal control).

SIMPLIFIED SENTENCE: 'The tests were Out of these, there were 39 cases with viral bronchiolitis from HPIV, 34 cases of non-tuberculosis mycobacteria (NTM), and 21 healthy people for comparison.

RESPONSE:
{{
 'No error': 0,
 'A1. Random generation': 0,
 'A2. Syntax error': 1,
 'A3. Contradiction': 0,
 'A4. Simple punctuation / grammar errors': 0,
 'A5. Redundancy': 0,
 'B1. Format misalignement': 0,
 'B2. Prompt misalignement': 0,
 'C1. Factuality hallucination': 0,
 'C2. Faithfulness hallucination': 0,
 'C3. Topic shift': 0,
 'D1.1. Overgeneralization': 0,
 'D1.2 Overspecification of Concepts': 0,
 'D2.1. Loss of Informative Content': 0,
 'D2.2. Out-of-Scope Generation': 0

}}

Now evaluate the next pair:

SOURCE SENTENCE:
{source_sentence}

SIMPLIFIED SENTENCE:
{simplified_sentence}

Do not include any text or commentary outside of the JSON response format below.
Respond in this JSON format:
{{
 'No error': float,
 'A1. Random generation': float,
 'A2. Syntax error': float,
 'A3. Contradiction': float,
 'A4. Simple punctuation / grammar errors': float,
 'A5. Redundancy': float,
 'B1. Format misalignement': float,
 'B2. Prompt misalignement': float,
 'C1. Factuality hallucination': float,
 'C2. Faithfulness hallucination': float,
 'C3. Topic shift': float,
 'D1.1. Overgeneralization': float,
 'D1.2 Overspecification of Concepts': float,
 'D2.1. Loss of Informative Content': float,
 'D2.2. Out-of-Scope Generation': float

}}
\end{lstlisting}

\end{document}

%% file: sections/00_main.tex
\section{Introduction} \label{introduction}

With the rise of social media's approach to learning via "quick bites", demand for lay-accessible content has surged. As a result, there is an increasing need for scientific information to be distilled into short, concise, and factually accurate content while remaining relatively simple. This combination of criteria is inherently challenging to satisfy, and there remain many disparate challenges to tackle, such as the risk of oversimplification, the potential for misinformation, and the loss of critical context.

Text Simplification has long been a goal in natural language processing, and has recently become a prominent task in machine learning research. Recognizing the core ideas of a paragraph, extracting the most pertinent information, and paraphrasing it is no small feat—even for humans. What poses an obstacle for machine learning techniques is the lack of clear measures for evaluating the "goodness" of a simplified text. Several metrics have been devised, but they mostly compare the simplification to a reference human summary, which is subjective in itself.

With the advent of LLMs, many NLP tasks have benefited, and text simplification is no exception. However, new problems naturally arose. One such problem is hallucinations, where an LLM generates spurious information without any basis in the reference or source text. The causes of hallucination in LLMs have been heavily researched, but remain unclear. This complicates mitigation endeavors, requiring considerable effort to prevent LLMs from producing misinformation or irrelevant content.

The SimpleText track of CLEF 2025 \cite{Ermakova2025Overview} has several tasks focusing on the simplifying scientific text problem. In this paper, we detail our submission to its second task, hallucination detection and mitigation in text simplification \cite{vendeville2025simpletext}. There are three subtasks:
\begin{itemize}
    \item \textbf{Task 2.1:} Identifying creative generation at document level
    \item \textbf{Task 2.2:} Detection and classification of information distortion errors in simplified sentences
    \item \textbf{Task 2.3:} Avoiding creative generation and performing grounded generation by design
\end{itemize}

In this paper, we present a unified system that tackles the three subtasks of hallucination detection using a combination of machine learning models and large language models (LLMs). For Tasks 2.1 and 2.2, we use multiple strategies to evaluate whether a simplified sentence is spurious or distorted, including a fine-tuned BERT classifier, a semantic similarity model that compares embeddings, an entailment model trained to detect contradictions, and an LLM prompted to act as a reasoning-based evaluator. The results from these different components are combined using a small neural network that learns how to make the final decision based on all available signals. For Task 2.3, which involves generating grounded simplifications, we prompt an LLM to revise simplified text using the original source as a reference, correcting any inaccurate or unrelated information. This layered setup allows us to catch different types of hallucinations and enforce factual consistency across tasks.

The paper is organized as follows: Section \ref{related work} reviews the state of the art in hallucination detection and mitigation in text simplification; Section \ref{methodology} outlines our approach to each subtask; Section \ref{results} presents and discusses the results; and Section \ref{conclusions} concludes the paper and discusses possible future directions.

\section{Related Work} \label{related work}

Hallucination in natural language generation (NLG) refers to the phenomenon where models produce content that is not supported or entailed by the input source \cite{ji2023survey}. This issue is particularly critical in scientific text simplification, where factual fidelity is essential. Several prior works have explored hallucination in the context of machine translation, summarization, and more recently, simplification.

Early efforts to detect hallucinations relied on heuristic-based methods, such as lexical overlap or n-gram precision metrics (e.g., BLEU), which have been shown to be inadequate for capturing semantic inconsistencies \cite{maynez2020faithfulness}. Recent advances incorporate entailment-based evaluation using natural language inference (NLI) models \cite{kryscinski2020evaluating}, which judge whether the generated text is logically supported by the source.

To mitigate hallucination in simplification, researchers have investigated controlled generation frameworks. These include constrained decoding \cite{dong2020multi}, post-editing using retrieval-based systems \cite{reid2022revisiting}, and reinforcement learning with factuality-based rewards \cite{gao2022factually}. 

In the context of scientific texts, hallucination is more severe due to the density and complexity of the source material. \citet{nishino2023scitldr} proposed SciTLDR, a dataset for extreme summarization of scientific papers, and highlighted hallucination issues in LLM outputs. Similarly, \citet{vendeville2025resource} introduced fine-grained annotations for information distortion errors in simplification, enabling targeted detection of hallucinations, overgeneralizations, and contradictions.

Large Language Models (LLMs), such as GPT and LLaMA, have shown strong performance in simplification tasks but remain prone to hallucination \cite{ziems2023can}. Techniques such as few-shot prompting, retrieval-augmented generation (RAG), and post-hoc correction using entailment scores have been used to reduce hallucinations in their output.

Our work builds on these foundations by combining multiple detection strategies. We further propose a grounded generation framework where LLMs act as faithful post-editors.

\section{Methodology} \label{methodology}

The focus of task 2 is on identifying and evaluating creative generation and information distortion in text simplification.

\subsection{Spurious Text Detection - Task 2.1}

This task requires identifying whether the input text is \textit{spurious} or not, both with access to its source abstract (subtask : \textbf{sourced}) and without access to the source (subtask : \textbf{post-hoc}). We apply an overall ensemble approach for both subtasks; however, the approaches diverge based on the availability of the source text.

The spurious detection problem is tackled by combining four complementary approaches into a unified framework. Each approach contributes distinct signals about the spuriousness of an input text. The outputs from these methods are further fused using an ensemble model:

\begin{enumerate}
    \item \textbf{BERT Classifier:} A fine-tuned BERT-based binary classification model trained to detect whether an input sentence is spurious or not. It captures lexical and syntactic cues through supervised learning.

    \item \textbf{Cosine Similarity Score:} A semantic similarity score computed using sentence embeddings. It measures how closely the input text aligns with the source in embedding space. Low similarity may indicate off-topic or fabricated content, while High similarity suggests grounding.

    \item \textbf{Pre-trained NLI Model:} A natural language inference (NLI) model is trained on entailment tasks. Given a reference text, It is used to assess whether the input text contradicts, entails, or is unrelated to the reference, providing a logic-based signal for spuriousness.

    \item \textbf{LLM as a Judge:} A large language model prompted with task-specific instructions to act as a reasoning-based evaluator. It reviews the given reference and input text and provides a verdict on whether the input contains hallucinations, contradictions, or irrelevant content [Appendix~\ref{app:prompt-LLM-as-Judge}].

    \item \textbf{Spurious Ensemble Detector:} A lightweight meta-classifier that takes as input the prediction scores from the BERT Classifier, Cosine similarity score, NLI model, and LLM Judge. It combines their outputs using a small neural network to make a final spuriousness prediction, leveraging the strengths of all three sources.
\end{enumerate}

\subsubsection{Sourced}

\begin{figure}[h]
    \centering
    \includegraphics[width=0.8\textwidth]{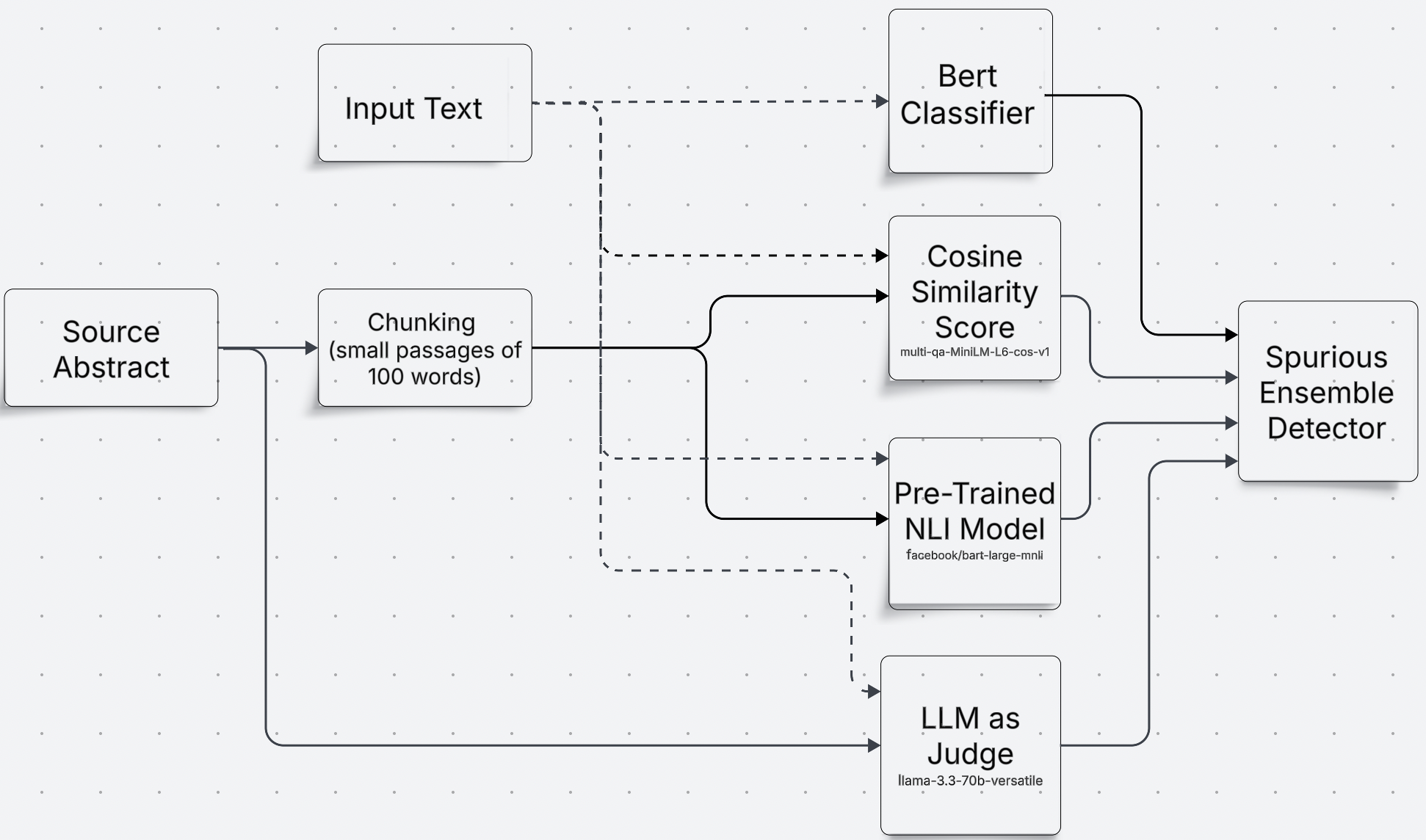} 
    \caption{Spurious Text Detection with access to Source - Task 2.1}
    \label{fig:task_2_1_sourced}
    \end{figure}

In the \textbf{sourced} setting, the input sentence is provided with a source abstract. The Bert Classifier is trained only on input text without using any source information. As such, it relies entirely on intrinsic textual features and learned patterns to make determinations about the spuriousness of the input text. To effectively utilize long-source abstracts, which often exceed standard input limits of transformer models, we adopt a document chunking strategy. Each source abstract is segmented into overlapping passages (or "chunks") of fixed length—specifically, 100 words per chunk with a 50-word overlap. This ensures coverage and contextual continuity across segments. Figure~\ref{fig:task_2_1_sourced} illustrates the architecture of our system in this setting, and different components are described below:

\begin{itemize}
    \item \textbf{Cosine Similarity Score:} For each input text, we compute the maximum cosine similarity between its embedding and those of all its source abstract chunks, using the Sentence Transformer model \texttt{multi-qa-MiniLM-L6-cos-v1}. This score captures the semantic alignment between the input and its source content.

    \item \textbf{Pre-trained NLI Model:} We utilize the natural language inference model \texttt{facebook/bart-large-mnli} to evaluate semantic consistency. For each chunk of the source abstract, we compute the entailment and contradiction probabilities with respect to the input text and retain the highest values across all chunks.

    \item \textbf{LLM as Judge:} A large language model, \texttt{llama-3.3-70b-versatile}, is prompted with both the source abstract and the input text using few-shot prompting. The model assigns scores in the range [0, 1] for four dimensions: \textit{spuriousness}, \textit{over-generalization}, \textit{contradiction}, and \textit{vagueness}, based on its reasoning capabilities.

    \item \textbf{Spurious Ensemble Detector:} A three-layer neural network classifier is trained to aggregate the eight probabilistic features derived from the previous approaches. This ensemble model predicts whether the input text is spurious by learning from complementary signals across bert-classifier, similarity, entailment, and judgment-based reasoning.

\end{itemize}

\subsubsection{Post-hoc}
\begin{figure}[h]
    \centering
    \includegraphics[width=0.8\textwidth]{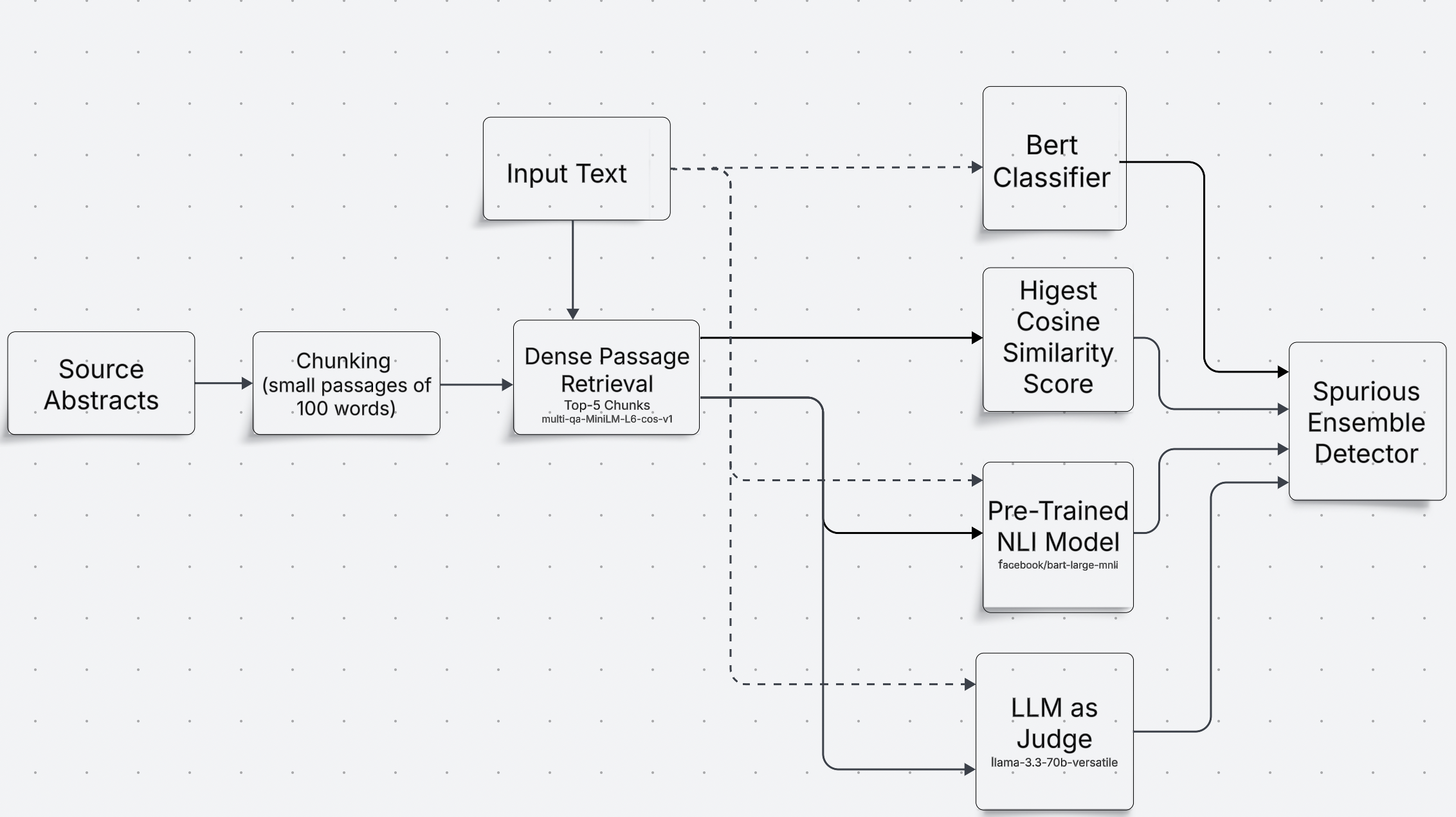} 
    \caption{Spurious Text Detection without access to Source - Task 2.1}
    \label{fig:task_2_1_posthoc}
    \end{figure}

The input texts are without access to their source in the \textbf{post-hoc} setting. There is no change in how the Bert Classifier is implemented compared to the sourced setting. The model is trained only on the input texts to determine if they are spurious or not. As illustrated in Figure~\ref{fig:task_2_1_posthoc}, the system architecture mirrors that of the sourced setting with some differences that are described below:

\begin{itemize}
    \item \textbf{Dense Passage Retrieval:} We utilize a pre-trained sentence embedding model, \texttt{multi-qa-MiniLM-L6-cos-v1}, to encode both the input text and the chunks derived from the source abstracts. Each source abstract is first segmented into 100-word chunks. These chunks are then embedded into a dense vector space. Given an input sentence, we compute its embedding and retrieve the top-5 most semantically similar chunks using cosine similarity. Let $q$ be the embedding of the input sentence and ${d_1, d_2, \dots, d_n}$ be the embeddings of source chunks. The top-$k$ chunks (here, $k=5$) with the highest cosine similarity scores are selected as the most relevant context.

    \item \textbf{Cosine Similarity Score:} For each input text, the highest cosine similarity score of the Dense Passage Retrieval is used.

    \item \textbf{Pre-trained NLI Model:} For the top-5 most semantically similar chunks retrieved, we compute the entailment and contradiction probabilities with respect to the input text and retain the highest values across all chunks.

    \item \textbf{LLM as Judge:} The top-5 highest cosine similar chunks retrieved for each input text are concatenated. A large language model, \texttt{llama-3.3-70b-versatile}, is prompted with concatenated source chunks and the input text using few-shot prompting.

     \item \textbf{Spurious Ensemble Detector:} Same as the \textbf{ source} setting, A three-layer neural network classifier is trained to aggregate the eight probabilistic features derived from previous approaches.

\end{itemize}

\subsection{Detecting and Classifying Information Distortion Errors - Task 2.2}
This task focuses on detecting information distortion in simplified sentences and classifying them into different types of errors \cite{vendeville2025resource}. We approached the multi-label classification problem through distinct strategies, leveraging both transformer-based models and large language models (LLMs).

\begin{figure}[h]
    \centering
    \includegraphics[width=0.8\textwidth]{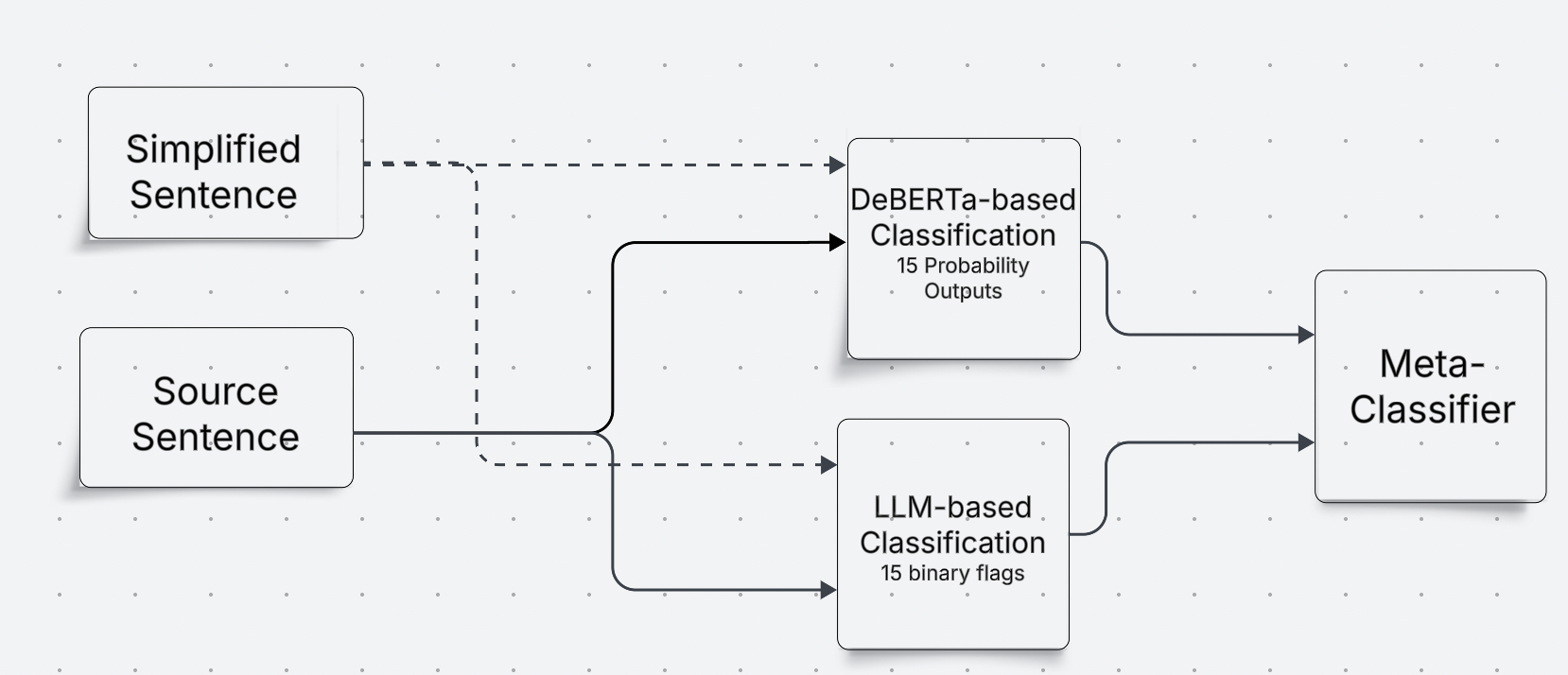} 
    \caption{Information Distortion Errors Classification in Simplified Sentences - Task 2.2}
    \label{fig:task2_2}
    \end{figure}

We explored fine-tuning a \texttt{RoBERTa-large} model for multi-label text classification. Similarly, we investigated LLM-based classification [Appendix~\ref{app:prompt-LLM Classifier}] using \texttt{LLaMA-3.3-70B-Versatile} to flag errors in the simplified sentence when compared to the source sentence. As illustrated in Figure~\ref{fig:task2_2}, We combined the strengths of both approaches through an ensemble framework, where the probability outputs from the DeBERTa model and the binary flags from the LLM are used as inputs to a three-layer neural network-based meta-classifier.

\subsection{Avoiding Creative Generation by performing Grounded Generation - Task 2.3}
This task requires a pair of Task 1 text simplification submissions, where one functions as the baseline approach and the other applies grounding to avoid overgeneration or other additional content not in the source documents or sentences.

\subsubsection{LLM-Based Grounded Generation}

To ensure factual consistency in the simplified text, we leverage \texttt{LLaMA-3.3-70B-Versatile} as a grounded generator. The model is prompted with both the baseline simplified text and a corresponding source text and tasked with producing a corrected version of the simplified text if necessary. The goal is to revise the baseline simplified text to eliminate any hallucinations, contradictions, fabricated content, or over generalizations, ensuring that the output remains strictly grounded in the provided reference.

We use a structured prompt to guide the LLM’s reasoning and generation:

\begin{quote}
\small
You are given: \\
- A reference document \\
- An input text that may contain errors such as fabricated content, contradictions, hallucinations or overgeneration. \\

Your task is to revise the input text so that it is fully grounded in the reference document. The corrected version must: \\
- Be factually consistent with the reference \\
- Avoid introducing unrelated or inaccurate information \\

Return only the corrected version of the input text if it is needed, otherwise return the same input text. \\

\textbf{Reference Document:} \{reference\_doc\} \\
\textbf{Input Text:} \{input\_text\} \\
\textbf{Corrected Text:}
\end{quote}

\begin{figure}[h]
    \centering
    \includegraphics[width=0.8\textwidth]{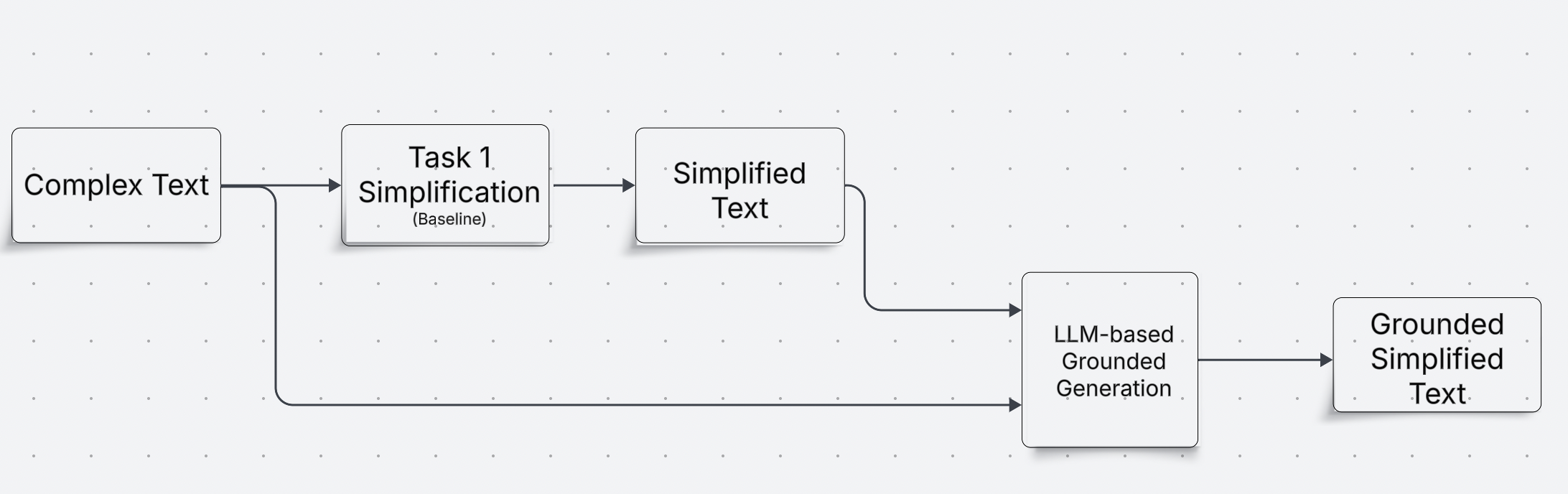} 
    \caption{LLM-Based Grounded Generation - Task 2.3}
    \label{fig:task2_3}
    \end{figure}

As depicted in Figure~\ref{fig:task2_3}, this prompting strategy enables the LLM to act as a post-editing agent that enforces alignment between the simplified text and the source text. When the baseline simplification is already grounded and accurate, the model returns it unchanged; otherwise, it produces a revised version that faithfully reflects the content and intent of the original reference.

\section{Results} \label{results}

\subsection{Evaluation of Spurious Text Detection - Task 2.1}
We evaluate the performance of our ensemble approach separately for both sub-tasks: \textbf{sourced}, where the input text is accompanied by its source abstract, and \textbf{post-hoc}, where the source is unavailable. In addition to reporting results for each setting independently, we also analyze whether the absence of source information in the post-hoc setting leads to any significant degradation in detection performance.

\subsubsection{Sourced}
To evaluate the effectiveness of our approach, we compare the performance of our ensemble model, which integrates predictions from the BERT Classifier, LLM as Judge, and NLI model, against each individual approach. Table~\ref{tab:task2.1-sourced} summarizes the evaluation metrics, including Accuracy, Precision, Recall, F1 Score, ROC AUC, AUPRC for this setting.

\begin{table}[h!]
\centering
\caption{Results for CLEF 2025 SimpleText Task 2.1 Detecting Overgeneration With Source Access}
\label{tab:task2.1-sourced}
\begin{tabular}{|l|r|c|c|c|c|c|c|}
\hline
\textbf{Method} & \textbf{Count} & \textbf{Accuracy} & \textbf{Precision} & \textbf{Recall} & \textbf{F1-score} & \textbf{AUROC} & \textbf{AUPRC} \\
\hline
\texttt{bertnlillmensemble} & 3,379 & 0.91 & 0.93 & 0.97 & 0.95 & 0.68 & 0.93 \\
\texttt{bertclassifier}     & 3,379 & 0.91 & 0.93 & 0.98 & 0.95 & 0.65 & 0.93 \\
\texttt{llm}                & 3,379 & 0.74 & 0.94 & 0.76 & 0.84 & 0.68 & 0.93 \\
\texttt{nli\_entailment}    & 3,379 & 0.35 & 0.92 & 0.31 & 0.46 & 0.53 & 0.90 \\
\texttt{nli\_contradiction} & 3,379 & 0.20 & 0.90 & 0.12 & 0.21 & 0.50 & 0.90 \\
\hline
\end{tabular}
\end{table}

\noindent While the BERT Classifier alone performs strongly with an F1 score of 0.95, the ensemble model offers a more balanced performance across metrics, especially in terms of precision and recall. The LLM as Judge provides high precision (0.94) but lower recall (0.76), suggesting conservative but accurate flagging of spurious text. The NLI model performs comparatively poorly in isolation, likely due to its sensitivity to phrasing and lack of contextual understanding. However, when combined in the ensemble, it contributes complementary signals, enhancing robustness.

Overall, the ensemble model achieves high performance (ROC AUC: 0.68) and demonstrates the effectiveness of combining shallow, semantic, and reasoning-based components for the spuriousness detection task.

\subsubsection{Post-hoc}
In the \textit{post-hoc} setting, the system must detect spurious content without access to the source abstract. We do a similar comparison of the performance of the ensemble model—combining BERT Classifier, LLM as Judge, and NLI Entailment model—with each individual component model. Table~\ref{tab:task2.1-posthoc} reports Accuracy, Precision, Recall, F1 Score, ROC AUC, AUPRC in this setting.

\begin{table}[h!]
\centering
\caption{Results for CLEF 2025 SimpleText Task 2.1 Detecting Overgeneration for Post-hoc without Source Access}
\label{tab:task2.1-posthoc}
\begin{tabular}{|l|r|c|c|c|c|c|c|}
\hline
\textbf{Method} & \textbf{Count} & \textbf{Accuracy} & \textbf{Precision} & \textbf{Recall} & \textbf{F1-score} & \textbf{AUROC} & \textbf{AUPRC} \\
\hline
\texttt{bertnlillmensemble} & 3,336 & 0.90 & 0.93 & 0.97 & 0.95 & 0.64 & 0.93 \\
\texttt{bertclassifier}     & 3,336 & 0.91 & 0.93 & 0.97 & 0.95 & 0.64 & 0.93 \\
\texttt{llm}                & 3,336 & 0.77 & 0.95 & 0.78 & 0.86 & 0.70 & 0.94 \\
\texttt{nli\_entailment}    & 3,336 & 0.45 & 0.95 & 0.41 & 0.57 & 0.61 & 0.92 \\
\hline
\end{tabular}
\end{table}

\noindent
The results show that the BERT Classifier alone performs very competitively, achieving the highest Accuracy (0.91), even slightly outperforming the ensemble model (0.90). This suggests that, in the absence of source information, the classifier trained purely on input text features is highly effective. The LLM as Judge exhibits high precision (0.95), indicating that it is conservative and reliable in flagging spurious text. However, it sacrifices recall (0.78), which reduces its overall F1 performance. While the ensemble approach does not outperform the BERT-only model in this post-hoc setting, it maintains robust and consistent performance by integrating diverse perspectives, making it resilient to individual model weaknesses.

\subsubsection{Overall Performance}
We observe that the performance of the ensemble model remains remarkably stable across both settings. The accuracy and F1 score show only a marginal decline when source information is unavailable (accuracy drops from 0.91 to 0.90), suggesting that the model is largely resilient to the absence of source grounding.

However, a more noticeable decline is observed in the \textbf{ROC AUC} metric, which drops from 0.68 in the sourced setting to 0.64 in the post-hoc setting. This suggests that while the model still performs well in binary classification, its ability to rank predictions with high confidence across the full score distribution is somewhat diminished without access to source context.

Overall, the results indicate that the absence of the source abstract does not significantly impair the model’s detection capabilities, thanks to the strong contribution of the BERT-based classifier and LLM judgment mechanisms.

\subsection{Evaluation of Information Distortion Error Classification - Task 2.2}
We evaluate the performance of the ensemble model that aggregates the probability outputs from DeBERTa-based classifier and the binary flags from the LLM to form a robust \textit{Meta-Classifier} for detecting various types of information distortion errors.

Table~\ref{tab:dsgt-error-performance} reports the F1 and AUC-PR scores of the ensemble model compared to individual classifiers, including RoBERTa-large, BERT, and LLM-based approaches.

\begin{table}[h!]
\centering
\caption{Model Performance by Error Categories for No error, Fluency(A), Alignment(B), Information (C), and Simplification (D) categories with F1 and AUC-PR}
\label{tab:dsgt-error-performance}
\begin{tabular}{|l|cc|cc|cc|cc|cc|}
\hline
\textbf{Method} & \multicolumn{2}{c|}{\textbf{No Error}} & \multicolumn{2}{c|}{\textbf{A}} & \multicolumn{2}{c|}{\textbf{B}} & \multicolumn{2}{c|}{\textbf{C}} & \multicolumn{2}{c|}{\textbf{D}} \\
 & F1 & AUC & F1 & AUC & F1 & AUC & F1 & AUC & F1 & AUC \\
\hline
DebertaLlmensemble & 0.763 & 0.561 & 0.283 & 0.133 & 0.354 & 0.173 & 0.301 & 0.156 & 0.374 & 0.224 \\
roberta            & 0.694 & 0.491 & 0.233 & 0.121 & 0.249 & 0.101 & 0.114 & 0.089 & 0.128 & 0.164 \\
llama              & 0.680 & 0.483 & 0.282 & 0.132 & 0.324 & 0.182 & 0.269 & 0.147 & 0.306 & 0.196 \\
BERT               & 0.515 & 0.330 & 0.214 & 0.133 & 0.208 & 0.103 & 0.167 & 0.095 & 0.129 & 0.161 \\
\hline
\end{tabular}
\end{table}

\noindent
The ensemble model consistently outperforms individual classifiers across all error categories. It achieves the highest F1 score (0.763) and AUC-PR (0.561) for correctly identifying instances with no errors, indicating strong precision and recall in distinguishing clean outputs. For all four error categories—Fluency (A), Alignment (B), Information (C), and Simplification (D)—the ensemble model yields superior F1 and AUC-PR scores compared to standalone models like \texttt{RoBERTa}, \texttt{LLaMA}, and \texttt{BERT}. 

While the LLaMA-based model shows competitive performance, particularly in Fluency (A) and Alignment (B), it falls short of the ensemble approach in detecting more nuanced errors such as Information loss (C) and Simplification issues (D). The BERT model, by contrast, lags in performance across all metrics, highlighting the advantage of using more advanced or combined architectures. 

Overall, these results demonstrate the benefit of leveraging an ensemble method that integrates DeBERTa and LLM-based predictions, especially in scenarios requiring nuanced error detection.

\subsection{Evaluation of Grounded Generation over Creative Generation - Task 2.3}

Table~\ref{tab:task2.3-dsgt} compares grounded submissions with baseline systems (denoted by $\star$) for CLEF 2025 SimpleText Task 2.3. Across both 37 aligned Cochrane-auto abstracts and 217 plain language summary test datasets, grounded systems generally exhibit higher semantic fidelity, as evidenced by consistently higher BLEU scores. For instance, \texttt{llama\_summary\_simplification\_grounded} outperforms the baseline with BLEU scores of 15.00 vs.\ 7.63 and 9.89 vs.\ 5.32. These results suggest that grounded simplification preserves the semantic content of the original text more faithfully. This is further supported by higher Levenshtein similarity scores and lower deletion proportions, indicating that grounded systems retain more of the original wording and structure.

However, the SARI scores—measuring the balance between addition, deletion, and copying operations for simplification—tend to be slightly lower for grounded models compared to their baseline counterparts. For example, \texttt{plan\_guided\_llama} (baseline) achieves a SARI of 42.98 versus 33.41 for its grounded variant on the 217 plain language summaries test dataset, suggesting that baseline models introduce more aggressive simplification operations.

This reflects a fundamental trade-off: grounded models, while producing more faithful and semantically aligned simplifications (as evidenced by higher BLEU and lexical similarity), may be less effective in performing bold rewrites or deletions that lead to simpler outputs, thus lowering their SARI scores. In summary, grounded systems are advantageous when semantic fidelity and contextual accuracy are prioritized.

\begin{table}[h!]
\centering
\caption{Results for CLEF 2025 SimpleText Task 2.3: Avoiding Creative Generation by Design}
\label{tab:task2.3-dsgt}
\resizebox{\textwidth}{!}{%
\begin{tabular}{|l|c|c|c|c|c|c|c|c|c|c|c|}
\hline
\textbf{Method} & \rotatebox{90}{\textbf{Count}} & \rotatebox{90}{\textbf{SARI}} & \rotatebox{90}{\textbf{BLEU}} & \rotatebox{90}{\textbf{FKGL}} & \rotatebox{90}{\textbf{Compression Ratio}} & \rotatebox{90}{\textbf{Sentence Splits}} & \rotatebox{90}{\textbf{Levenshtein Similarity}} & \rotatebox{90}{\textbf{Exact Copies}} & \rotatebox{90}{\textbf{Additions Proportion}} & \rotatebox{90}{\textbf{Deletions Proportion}} & \rotatebox{90}{\textbf{Lexical Complexity Score}} \\
\hline
llama\_summary\_simplification\_grounded & 37 & 41.25 & 15.00 & 12.74 & 0.76 & 0.85 & 0.57 & 0.00 & 0.23 & 0.48 & 8.76\\
$\star$llama\_summary\_simplification & 37 & 40.32 & 7.63 & 9.56 & 0.59 & 0.86 & 0.42 & 0.00 & 0.31 & 0.70 & 8.49\\
plan\_guided\_llama\_grounded & 37 & 37.33 & 18.27 & 12.87 & 0.91 & 1.09 & 0.71 & 0.00 & 0.18 & 0.31 & 8.79\\
$\star$plan\_guided\_llama & 37 & 42.33 & 10.43 & 7.77 & 0.48 & 0.97 & 0.47 & 0.00 & 0.18 & 0.70 & 8.52\\
\hline
llama\_summary\_simplification\_grounded & 217 & 42.06 & 9.89 & 12.81 & 0.62 & 0.72 & 0.50 & 0.00 & 0.19 & 0.59 & 8.55\\
$\star$llama\_summary\_simplification & 217 & 42.92 & 5.32 & 9.94 & 0.49 & 0.72 & 0.39 & 0.00 & 0.24 & 0.75 & 8.82\\
plan\_guided\_llama\_grounded & 217 & 33.41 & 10.04 & 12.96 & 0.96 & 1.14 & 0.69 & 0.00 & 0.21 & 0.31 & 8.88\\
$\star$plan\_guided\_llama & 217 & 42.98 & 6.33 & 7.82 & 0.48 & 0.99 & 0.46 & 0.00 & 0.18 & 0.71 & 8.50\\
\hline
\end{tabular}
} 
\end{table}

\noindent





\section{Conclusions} \label{conclusions}
Large Language Models (LLMs) have demonstrated strong capabilities in the domain of scientific text simplification. However, they are also prone to hallucinations and overgeneration, which can compromise the faithfulness and factual accuracy of the generated content.

In this work, we proposed ensemble-based approaches to identify spurious content and information distortion errors in simplified text. Our methods combine the strengths of individual components—including BERT-based classifiers, NLI models and LLM-based classifiers and together form a robust detection framework. For grounded simplification, we further demonstrated that LLMs serve effectively as high-precision post-editors, capable of revising simplified text to maintain consistency with the source document while correcting factual errors.

Our experiments show that spurious text detection degrades only slightly in post-hoc settings where the source document is unavailable. However, a notable drop in ROC AUC suggests that additional investigation is needed to better understand this discrepancy. Furthermore, in comparing grounded generation with baseline simplification, we observe a clear trade-off between simplification and faithfulness. Grounded outputs are more accurate but often more complex. 

Future work will explore techniques to optimize this trade-off, aiming to preserve simplicity without sacrificing factual correctness or introducing hallucinations.

\section*{Acknowledgements}

We thank the Data Science at Georgia Tech (DS@GT) CLEF competition group for their support.
This research was supported in part through research cyberinfrastructure resources and services provided by the Partnership for an Advanced Computing Environment (PACE) at the Georgia Institute of Technology, Atlanta, Georgia, USA \cite{PACE}. 

\section*{Declaration on Generative AI}

 During the preparation of this work, the authors used ChatGPT and Gemini for grammar and spelling check, as well as assistance in the code for the conducted experiments.  After using these tools, the authors reviewed and edited the content as needed and take full responsibility for the publication’s content. 